\pdfoutput=1

\documentclass[11pt]{article}

\usepackage{acl}

\usepackage{times}
\usepackage{latexsym}

\usepackage[T1]{fontenc}

\usepackage[utf8]{inputenc}

\usepackage{microtype}

\usepackage{inconsolata}

\usepackage{dirtytalk}

\usepackage{graphicx}
\usepackage{todonotes}
\usepackage{url}
\usepackage{arydshln}
\usepackage{multirow}
\usepackage{adjustbox}
\usepackage{amsmath}
\usepackage{booktabs}
\usepackage[framemethod=TikZ]{mdframed}

\makeatletter
\def\adl@drawiv#1#2#3{%
        \hskip.5\tabcolsep
        \xleaders#3{#2.5\@tempdimb #1{1}#2.5\@tempdimb}%
                #2\z@ plus1fil minus1fil\relax
        \hskip.5\tabcolsep}
\newcommand{\cdashlinelr}[1]{%
  \noalign{\vskip\aboverulesep
           \global\let\@dashdrawstore\adl@draw
           \global\let\adl@draw\adl@drawiv}
  \cdashline{#1}
  \noalign{\global\let\adl@draw\@dashdrawstore
           \vskip\belowrulesep}}
\makeatother

\newcommand{\quotes}[1]{``#1''}
\renewcommand{\vec}[1]{\ensuremath{\mathbf{#1}}}

\definecolor{lightgray}{rgb}{0.97,0.97,0.97} 
\definecolor{darkgray}{rgb}{0.8,0.8,0.8} 

\author{
 \textbf{Jakub \v{S}m\'{i}d\textsuperscript{*}},
 \textbf{Pavel P\v{r}ib\'{a}\v{n}\textsuperscript{*}},
 \textbf{Pavel Kr\'{a}l\textsuperscript{*, $\dagger$}}
\\
\\
University of West Bohemia, Faculty of Applied Sciences, \\
 \textsuperscript{*}Department of Computer Science and Engineering,\\
 \textsuperscript{$\dagger$}NTIS – New Technologies for the Information Society\\
         Univerzitní 2732/8, 301 00 Pilsen, Czech Republic \\
    \{jaksmid, pribanp, pkral\}@kiv.zcu.cz\\
         \url{https://nlp.kiv.zcu.cz}
}

\usepackage{fancyhdr}

\fancypagestyle{firstpage}{
  \fancyhf{} 
  \fancyfoot[C]{\thepage} 
  \fancyfoot[L]{\scriptsize Accepted at WASSA 2024. Official version: \href{https://doi.org/10.18653/v1/2024.wassa-1.47}{10.18653/v1/2024.wassa-1.47}} 
}

\title{UWB at WASSA-2024 Shared Task 2: Cross-lingual Emotion Detection}

\begin{document}
\maketitle
\thispagestyle{firstpage} 

\begin{abstract}
This paper presents our system built for the WASSA-2024 Cross-lingual Emotion Detection Shared Task. The task consists of two subtasks: first, to assess an emotion label from six possible classes for a given tweet in one of five languages, and second, to predict words triggering the detected emotions in binary and numerical formats. Our proposed approach revolves around fine-tuning quantized large language models, specifically Orca~2, with low-rank adapters (LoRA) and multilingual Transformer-based models, such as XLM-R and mT5. We enhance performance through machine translation for both subtasks and trigger word switching for the second subtask. The system achieves excellent performance, ranking 1st in numerical trigger words detection, 3rd in binary trigger words detection, and 7th in emotion detection.
\end{abstract}

\section{Introduction}
Analyzing emotions in text, including emotion detection and other related tasks, is a well-studied area in natural language processing (NLP). This field has been extensively explored in various SemEval~\citep{mohammad-etal-2018-semeval, chatterjee-etal-2019-semeval} and WASSA~\citep{klinger-etal-2018-iest} competitions. The goal of WASSA-2024 Shared Task 2~\citep{Maladry2024} is to predict specific emotions and identify the words that trigger these emotions. Additionally, this study investigates how emotional information can be transferred across five languages. While the training data is provided in English, the evaluation data includes English, Dutch, Russian, Spanish, and French.

Specifically, the task consists of two subtasks: \\
\hspace*{5pt} 1) Cross-lingual emotion detection: Predicting emotion from six possible classes (\textit{Love}, \textit{Joy}, \textit{Fear}, \textit{Anger}, \textit{Sadness}, \textit{Neutral}) in five target languages. \\
\hspace*{5pt} 2) Classifying words that express emotions: Identifying words that trigger emotions, with the output format being either binary (assigning a~binary value to each token in the text) or numeric (assigning a numeric value to each token in the text). 

\begin{figure}[ht!]
    \centering
    \includegraphics[scale=0.7]{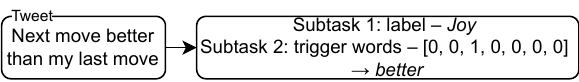}
    \caption{Example tweet with labels for both subtasks.}
    \label{fig:task-example}
\end{figure}

Figure~\ref{fig:task-example} shows an example from the dataset for both subtasks. For detailed label descriptions, see the annotation guidelines~\citep{01HHSH3CDPHWA217Y73PHAVK5R}.

The first subtask can be considered a sentence-level classification task, while the second can be treated as a token-level classification task. Both subtasks can also be viewed as text-generation tasks~\citep{raffel2023exploring}. Learning the representation of a given text is essential for solving these problems. Neural networks, including convolutional neural networks (CNNs)~\citep{kim-2014-convolutional} and recurrent neural networks (RNNs)~\citep{zaremba2015recurrent}, alongside their variations, have been employed for this purpose. However, while these models are effective, they primarily rely on static word embeddings to capture semantic meanings. Consequently, they may struggle with more complex linguistic features such as anaphora and long-term dependencies.

Recent NLP research has shifted towards Transformer-based pre-trained language models (PLMs), such as BERT~\citep{devlin-etal-2019-bert} and T5~\citep{raffel2023exploring}. These models undergo initial pre-training on extensive datasets to grasp language representation intricacies. Subsequently, they can be fine-tuned on labelled data, capitalizing on the knowledge acquired during pre-training. For cross-lingual tasks, multilingual pre-trained models such as mT5~\citep{xue-etal-2021-mt5} and XLM-RoBERTa (XLM-R)~\citep{conneau-etal-2020-unsupervised} have emerged as standard choices~\citep{pmlr-v119-hu20b}.

Recently, open-source large language models (LLMs), such as LLaMA 2~\citep{touvron2023llama} and Orca~2~\citep{mitra2023orca}, have made significant progress across various NLP tasks. These models show remarkable performance in many zero- and few-shot tasks. Nevertheless, they are primarily pre-trained on English, which often necessitates additional fine-tuning to optimize their performance for other languages~\citep{zhang-etal-2023-dont}. However, fine-tuning LLMs on non-specialized consumer GPUs presents challenges due to their large number of parameters. Techniques like QLoRA~\citep{dettmers2023qlora} address this issue by employing a quantized 4-bit frozen backbone LLM with a small set of learnable LoRA weights~\citep{hu2021lora}.

This paper proposes improving cross-lingual emotion detection by combining a quantized Orca~2 LLM, fine-tuned with LoRA, and machine translation. Additionally, we leverage fine-tuned Transformer-based multilingual language models, such as XLM-R and mT5, for trigger word detection. By incorporating alignment-free translation and trigger word switching, we aim to enhance performance further.\footnote{The code is publicly available at \url{https://github.com/biba10/UWBWASSA2024SharedTask2}.}

\section{System Description}
We conduct experiments using the dataset provided by \citet{Maladry2024}, which contains tweets in five different languages.

\subsection{Problem Formulation}
For both subtasks, the input is a sentence $x = \{x_i\}_{i=1}^L$ containing $L$ tokens. We denote the parameters of the models as $\vec{\Theta}$, which includes task-specific parameters $\vec{W}$ and $\vec{b}$. Given the sentence-label pairs $(x^S, y^S)$ in the source language $S$, the aim is to predict labels $y^T$ for the sentence $x^T$ in the target language $T$.

We formulate the emotion detection subtask as a text-generation problem, which can be modelled, for example, with decoder-only Transformer-based models. The decoder-only model calculates the probability of generating the next token $y_t$ at each step $t$ based on previous outputs $y_1, \ldots, y_{t-1}$ as
\begin{equation}
    P_\vec{\Theta}(y_t|y_1, \ldots, y_{t-1}) = \mathbf{Dec}(y_1, \ldots, y_{t-1}),
\end{equation}
where $\mathbf{Dec}$ is the decoder function.

We formulate the second subtask as a token-level classification problem. Given the feature representations $\vec{h} = \{\vec{h}\}_{i=1}^L$ for each token in the sentence obtained by the model, we apply a linear layer on top to get the label distribution for each token $x_i$ as
\begin{equation}
    P_\vec{\Theta}(y_i|x_i) = \operatorname{softmax}(\vec{W}\vec{h}_i + \vec{b}),
\end{equation}
where $y \in \mathcal{Y} = \{0, 1\}$. We select the class with the highest probability. 
To obtain the numerical values, we extract the logits for class 1, which represents \textit{\quotes{is a trigger word}} class, and apply the softmax function to these logits to get numbers between 0 and 1. We consider only the first token of each word for both binary and numerical values.

\subsection{Label Projection}
Following related work in cross-lingual classification~\citep{hassan-etal-2022-cross, zhang-etal-2021-cross}, we translate the English training set into all non-English target languages using the Google API\footnote{\url{https://translate.google.com/}}. This approach significantly expands the training data. The process for the emotion detection task is straightforward: we translate the data and retain the original labels.

\begin{figure*}[ht!]
    \centering
    \includegraphics[scale=0.8]{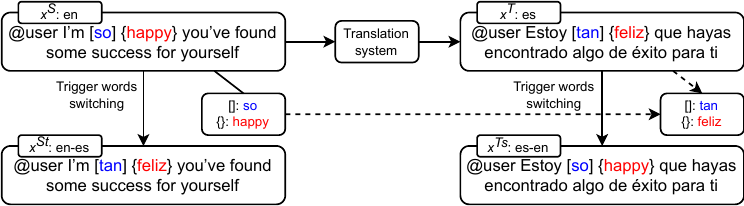}
    \caption{Example of label projection method with trigger words switching (lower part) for English and Spanish language pair.}
    \label{fig:tr}
\end{figure*}

Creating translated data for the second subtask is more challenging because the translated text may not perfectly align with the original English text, resulting in different token counts between the two.
Previous approaches~\citep{mayhew-etal-2017-cheap, fei-etal-2020-cross} often rely on word alignment tools like fastAlign~\citep{dyer-etal-2013-simple} to map token-label pairs from the source sentence to the targeted one. We adopt an alignment-free label projection proposed by \citet{zhang-etal-2021-cross} to create pseudo-labelled data in the target languages.

First, we mark each trigger word in a sentence with predefined special symbols, such as \quotes{[]} or \quotes{\{\}}, before translating it. We use distinct symbols for a sentence containing multiple trigger words. After translation, we extract the trigger words from the translated sentence using the special symbols. The translation system may occasionally overlook these special symbols, leading to their omission. In such cases, we discard the sentences. This process, illustrated in Figure~\ref{fig:tr}, yields pseudo-labelled sentences in the target language.

Furthermore, we create additional datasets for the second subtask by combining data from different languages, as shown in the lower part of Figure~\ref{fig:tr}. Given an English source sentence $x^S$ and its translation $x^T$, we switch the trigger words to construct two new bilingual sentences: the first, $x^{St}$, originates from $x^S$ with trigger words expressed in the target language; the second, $x^{Ts}$, originates from $x^T$ with trigger words in the source language.

We denote the original English dataset as $D_S$, the translated dataset into all four non-English languages as $D_T$, the English source dataset with trigger words in other languages as $D_{St}$, and the translated dataset with trigger words in English as $D_{Ts}$.

\subsection{LoRA}
Fine-tuning LLMs like Orca~2 requires significant computational resources due to the model's extensive parameter count. To address this challenge, \citet{dettmers2023qlora} propose to use low-rank adapters (LoRA)~\citep{hu2021lora} on top of a quantized backbone model. This method employs a small set of trainable parameters called adapters while keeping the original model frozen, thus reducing memory requirements.

\begin{figure}[ht!]
    \centering
    \includegraphics[scale=0.37]{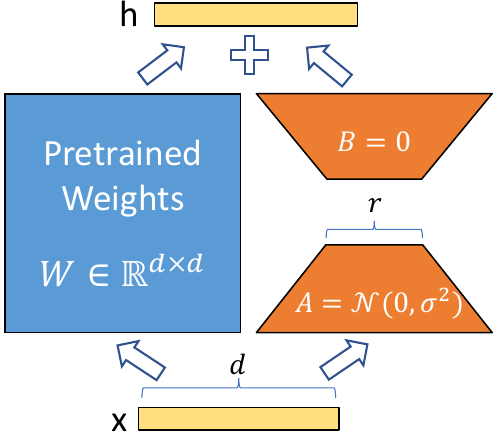}
    \caption{The illustrative depiction of parameter-saving LoRA fine-tuning proposed by \citet{hu2021lora}.}
    \label{fig:lora}
\end{figure}

For a pre-trained weight matrix $\vec{W}_0$, LoRA represents it with a low-rank decomposition as 
\begin{equation}
    \vec{W}_0 + \Delta\vec{W} = \vec{W}_0 + \vec{B}\vec{A},
\end{equation}
where $\vec{B}$ and $\vec{A}$ are matrices with much lower dimensions than $\vec{W}_0$.
During fine-tuning, $\vec{W}_0$ is frozen while the weights of $\vec{A}$ and $\vec{B}$ matrices are updated. Figure~\ref{fig:lora} shows the concept of LoRA.

\subsection{Models}
We fine-tune the large versions of XLM-R~\citep{conneau-etal-2020-unsupervised} and mT5~\citep{xue-etal-2021-mt5}, as well as the 13B version of Orca~2~\citep{mitra2023orca}, using the HuggingFace Transformers library\footnote{\url{https://github.com/huggingface/transformers}}~\citep{wolf-etal-2020-transformers}. The Orca~2 model is used for the first subtask, while the XLM-R model and the encoder part of the mT5 model are employed for the second subtask.

\subsection{Experimental Setup}
For the initial hyperparameter selection, we sample 10\% of the English training dataset as validation data for all experiments, following other cross-lingual work~\citep{hu-etal-2019-open, zhang-etal-2021-cross} that simulate true unsupervised settings. This 10\% is also excluded from the translation process that creates additional datasets. We use the original validation data in five languages to further reduce the number of models. 

For both subtasks, we always fine-tune the models on the original English dataset $D_S$ (excluding the 10\% of data used for validation) and explore the impact of incorporating additional data. Specifically, for the first subtask, we experiment with adding the translated dataset $D_T$. For the second subtask, we explore the following options: adding the translated dataset $D_T$, datasets with switched trigger words ($D_{St}$ and $D_{Ts}$), or both.

Figure~\ref{fig:prompt} shows the prompt used for fine-tuning the Orca~2 model, where we train the model to output the specific emotion class in a textual format.

\begin{figure}[ht!]
    \centering        
    \begin{mdframed}[backgroundcolor=lightgray, roundcorner=10pt, linewidth=1pt,
        innerleftmargin=3pt,
        innerrightmargin=3pt,
        innertopmargin=3pt,
        innerbottommargin=3pt,
        ]
                \setlength{\parindent}{0pt}
        \footnotesize
 Predict one emotion label for the given text. The possible labels are: \quotes{Love}, \quotes{Joy}, \quotes{Anger}, \quotes{Fear}, \quotes{Sadness}, \quotes{Neutral}.

 Answer in one following format: \quotes{Label: <emotion\_label>}
            \end{mdframed}
    \caption{Prompt for the emotion detection task.}
    \label{fig:prompt}
\end{figure}

\subsubsection{Hyperparameters}
We employ the AdamW optimizer~\citep{loshchilov2019decoupled} with a batch size 16 for all models. For Orca~2, we use QLoRA for fine-tuning with 4-bit quantization, setting LoRA $r=64$ and $\alpha=16$, a learning rate 2e-4 with linear decay, and applying LoRA adapters on all linear Transformer block layers. We fine-tune the model for up to 5 epochs. For other models, we fine-tune them for up to 30 epochs and search for the learning rate from \{2e-6, 2e-5, 5e-5, 2e-4\} using a constant scheduler. All experiments are conducted on an NVIDIA A40 with 48 GB GPU memory.

\subsubsection{Evaluation Metrics}
The main metric for the emotion detection subtask is the macro-averaged F1 score. The primary metric for the binary trigger word detection subtask is the token-level F1 score, calculated on a token level and averaged across instances. A new metric called accumulated precise importance attribution is used for the numerical trigger word detection subtasks. After normalization (ignoring negative values and ensuring the attributions for each sentence add up to 1), this metric sums up the attributions for each trigger word (i.e., the tokens with a label 1). 

\section{Results}
This section presents the results.

\subsection{Emotion Detection}
Table~\ref{tab:res1} shows the results of the emotion detection subtask. The results indicate that the Orca~2 largely benefits from the additional data translated into target languages, improving the results by over 3\% on test data and by more than 7\% on the validation set. The best model achieves a test score of 59.10, ranking seventh in the competition.

\begin{table}[ht!]
    \centering
    \begin{adjustbox}{width=0.6\linewidth}
        \begin{tabular}{@{}lrrc@{}}
            \toprule
            \textbf{Dataset} & \textbf{Dev}   & \textbf{Test} & \textbf{Rank} \\ \midrule
            $D_S$      & 50.12 &  55.73  &  \\
            $D_S$ + $D_T$ & 57.74 &  \textbf{59.10} & 7.   \\ 
            \cdashlinelr{1-4}
            Baseline & & 44.76 & \\
            Best &  & 62.95 & \\
            \bottomrule
        \end{tabular}
    \end{adjustbox}
    \caption{F1 macro scores on the emotion detection task with the Orca~2 model compared to baseline and best results~\citep{Maladry2024}. \textbf{Bold} indicates the officially announced results and their competition rank.}
    \label{tab:res1}
\end{table}

\begin{figure}[ht!]
    \centering
    \includegraphics[scale=0.53]{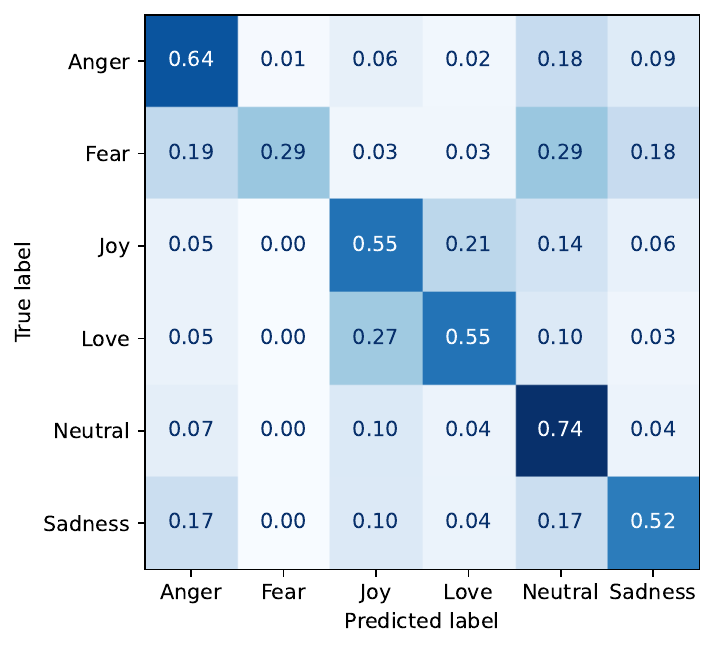}
    \caption{Confusion matrix on test data for the Orca~2 model fine-tuned on English data only.}
    \label{fig:confusion_matrix_en}
\end{figure}

\begin{figure}[ht!]
    \centering
    \includegraphics[scale=0.53]{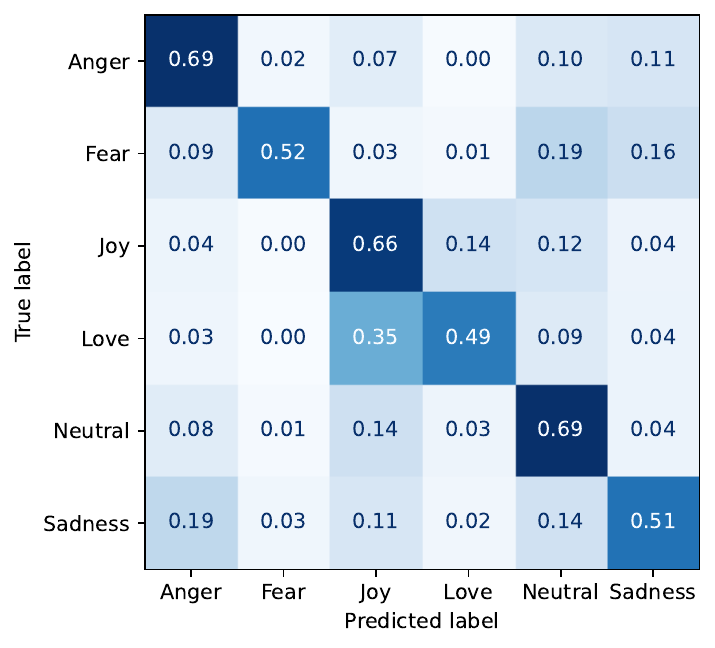}
    \caption{Confusion matrix on test data for the Orca~2 model fine-tuned on English and translated data.}
    \label{fig:confusion_matrix_tr}
\end{figure}

Figure~\ref{fig:confusion_matrix_en} and Figure~\ref{fig:confusion_matrix_tr} show the confusion matrices for the Orca~2 model fine-tuned on English data only and both English and translated data, respectively. The model fine-tuned on translated data demonstrates significantly better performance for the \textit{Fear} (0.52 vs 0.29) and \textit{Joy} (0.66 vs 0.55) classes while maintaining similar performance for other classes compared to the model fine-tuned only on English data.

Misclassified labels tend to cluster by sentiment. For instance, the \textit{Love} label is frequently misclassified as \textit{Joy}. The \textit{Neutral}, \textit{Anger}, and \textit{Joy} classes appear to be the easiest to predict, likely due to their higher representation in the training and test data, as shown in Table~\ref{tab:data}.

\begin{table}[ht!]
    \centering
    \small
    \begin{tabular}{@{}lrrr@{}}
        \toprule
        \textbf{Label} & \textbf{Train} & \textbf{Dev} & \textbf{Test} \\ \midrule
        Anger          & 1,028          & 129          & 614           \\
        Fear           & 143            & 14           & 77            \\
        Joy            & 1,293          & 102          & 433           \\
        Love           & 579            & 40           & 190           \\
        Neutral        & 1,397          & 157          & 916           \\
        Sadness        & 560            & 58           & 270           \\ \bottomrule
    \end{tabular}
\caption{Label distribution for the emotion detection.}
\label{tab:data}
\end{table}

\subsection{Trigger Words Detection}
Table~\ref{tab:res2} presents the results of the trigger words detection subtask, evaluating the performance for binary and numerical outputs using various training combinations. For binary classification, XLM-R achieves the highest test score (59.19) when trained on the full combination of datasets ($D_S + D_T + D_{St} + D_{Ts}$), ranking third in the competition. Conversely, mT5 excels in numerical value prediction with a top test score of 70.52 under the same training conditions, securing first place. In most cases, the mT5 model's overall performance on numerical triggers is substantially worse than the performance of XLM-R. However, the results improve significantly when all dataset combinations are used.

Determining the best dataset combination and model is challenging due to the similar results achieved across many cases. Variability from random seeds during fine-tuning can obscure slight performance differences, which may not truly indicate superiority but rather random fluctuations. Nonetheless, our best test set results are obtained by training on a combination of the original, translated, and trigger words switched datasets. Combining all datasets for numerical trigger word detection using the mT5 model significantly improves the second-best settings by 8\%.

\begin{table}[ht!]
    \centering
    \begin{adjustbox}{width=\linewidth}
        \begin{tabular}{@{}llrrcrrc@{}}
            \toprule
            \multirow{2}{*}{\textbf{Model}} & \multirow{2}{*}{\textbf{Dataset}} & \multicolumn{3}{c}{\textbf{Binary}} & \multicolumn{3}{c}{\textbf{Numerical}} \\ \cmidrule(lr){3-5} \cmidrule(lr){6-8}
                                            &                                   & Dev               & Test  & Rank          & Dev                & Test  & Rank            \\ \midrule
            \multirow{4}{*}{XLM-R}          & {\small $D_S$}                                & 61.85             &    58.59   &          & 70.07              &      70.06            & \\
                                            & {\small $D_S$ + $D_T$}                           & 60.82              &  56.69  &             & 71.52              &   66.20             &   \\
                                            & {\small $D_S$ + $D_{St}$ + $D_{Ts}$}                    & 57.97             &   53.18     &         & 73.14              &   70.16             &   \\
                                            & {\small $D_S$ + $D_T$ + $D_{St}$ + $D_{Ts}$}               & 60.46             &  \textbf{59.19} & 3.              & 70.05              &  70.02                & \\ \cdashlinelr{1-8} 
            \multirow{4}{*}{mT5}            & {\small $D_S$}                                & 59.99             &         58.12   &     & 60.11              &          60.00       &  \\
                                            & {\small $D_S$ + $D_T$}                           & 62.12             &   58.06          &    & 63.14              &  62.06               &  \\
                                            & {\small $D_S$ + $D_{St}$ + $D_{Ts}$}                 &   53.32                &   48.61    &          &  66.18                  &   62.86      &          \\
                                            & {\small $D_S$ + $D_T$ + $D_{St}$ + $D_{Ts}$}               & 59.22             & 56.79    &            & 70.92              &   \textbf{70.52} & 1.                \\ 
                                            \cdashlinelr{1-8} 
        Baseline & & & 23.49 & & & 21.60 & \\
        Best & & & 61.58 & & & 70.52 & \\
                                            \bottomrule 
        \end{tabular}
    \end{adjustbox}
    \caption{Token-level F1 score for binary trigger words detection and accumulate precise importance for numerical trigger words detection compared to baseline and best results~\citep{Maladry2024}. \textbf{Bold} indicates the officially announced results and their competition rank.}
    \label{tab:res2}
\end{table}

\subsection{Discussion}
Overall, the Orca~2 benefits more from the translated data than the XLM-R and mT5 models, likely because the Orca~2 is pre-trained mostly on English data. In contrast, the other two models are multilingual. The translated data adds more knowledge to a model pre-trained mainly on English data than those already exposed to multiple languages. In addition, trigger word detection tasks may be more prone to translation errors, which could diminish the benefits of translated data for these tasks. However, the mT5 model shows marginal improvements from the translated and trigger words switched datasets for the numerical trigger words detection, improving the results by 8\% and achieving the best result overall in the competition.

\section{Conclusion}
This paper describes our system for the WASSA-2024 Cross-lingual Emotion Detection Shared Task. We propose fine-tuning a quantized large language model with low-rank adapters combined with machine translation for the emotion detection subtask and fine-tuning multilingual Transformer-based models enhanced with machine translation and trigger word switching for the trigger words detection subtask. We show that additional translated data improves the performance. Our system achieves excellent results and ranks first in numerical trigger word detection, third in binary trigger word detection, and seventh in emotion detection.

\section*{Acknowledgements}
This work has been partly supported by the OP JAC project DigiTech no. CZ.02.01.01/00/23\_021/0008402 and by the Grant No. SGS-2022-016 Advanced methods of data processing and analysis.
Computational resources were provided by the e-INFRA CZ project (ID:90254), supported by the Ministry of Education, Youth and Sports of the Czech Republic.

\section*{Limitations}
The method relies on machine translation, thus its effectiveness depends on translation quality. In addition, we solve each subtask independently. Future research could address solving both subtasks simultaneously, potentially leading to a more robust model that better explains its decisions.

\section*{Ethics Statement}
We ensure fair and honest analysis while conducting our work ethically and without harming anybody. However, we acknowledge that the language models used in this paper may introduce unintended biases related to race or gender due to pre-training on large corpora.

\bibliography{anthology, bibliography}

\end{document}